\documentclass[conference,a4paper]{IEEEtran}
\IEEEoverridecommandlockouts

\usepackage[pagebackref,breaklinks,colorlinks]{hyperref}
\usepackage[cmex10]{amsmath}%American Math Society(AMS) math formatting
\usepackage{amssymb,amsfonts}%AMS extra symbols and fonts
\usepackage{dblfloatfix}%fix double column figure ordering and placement

\usepackage[ruled,vlined]{algorithm2e}
\usepackage{graphicx}
\graphicspath{{Figures/PDF/}{Figures/PNG/}}

\usepackage{booktabs}   % 用于三线表 (\toprule, \midrule, \bottomrule)
\usepackage{multirow}   % 用于跨行单元格 (\multirow)
\usepackage[table]{xcolor} % 用于表格颜色 (\rowcolor, \color)
\usepackage{graphicx}   % 用于调整大小 (\resizebox)

\usepackage{booktabs}
\usepackage{siunitx}
\usepackage[numbers,compress]{natbib}
\usepackage{texnames}
\usepackage{bm,bbm}
\usepackage{orcidlink}
\usepackage{dsfont}
\begin{document}

\title{\uppercase{GEOREASON: ALIGNING THINKING AND ANSWERING IN REMOTE SENSING VISION-LANGUAGE MODELS VIA LOGICAL CONSISTENCY REINFORCEMENT LEARNING}
\thanks{\IEEEauthorrefmark{2} The authors contribute equally to this work}
\thanks{\IEEEauthorrefmark{3} Corresponding author}
}

\author{
    % 1. 所有作者姓名放在同一个 IEEEauthorblockN 中，使用逗号或 and 连接
    \IEEEauthorblockN{
        Wenshuai Li\textsuperscript{1,}\textsuperscript{2,}\textsuperscript{3}\IEEEauthorrefmark{2}, 
        Xiantai Xiang\textsuperscript{1,}\textsuperscript{2,}\textsuperscript{3}\IEEEauthorrefmark{2},
        Zixiao Wen\textsuperscript{1,}\textsuperscript{2,}\textsuperscript{3},
        Guangyao Zhou\textsuperscript{1,}\textsuperscript{2,}\textsuperscript{3},
        Ben Niu\textsuperscript{1,}\textsuperscript{2}\IEEEauthorrefmark{3}\\
        Feng Wang\textsuperscript{1,}\textsuperscript{2},
        Lijia Huang\textsuperscript{1,}\textsuperscript{2,}\textsuperscript{3},
        Qiantong Wang\textsuperscript{1,}\textsuperscript{2},
        Yuxin Hu\textsuperscript{1,}\textsuperscript{2,}\textsuperscript{3}
    }
    
    \IEEEauthorblockA{
        \textit{$^1$Aerospace Information Research Institute, Chinese Academy of Sciences}\\
        \textit{$^2$Key Laboratory of Target Cognition and Application Technology, Chinese Academy of Sciences}\\
        \textit{$^3$University of Chinese Academy of Sciences}
    }
    \IEEEauthorblockA{Code: \url{https://github.com/canlanqianyan/GeoReason}}
}
\maketitle
\vspace{-20pt}
\begin{abstract}
% The advancement of Remote Sensing Vision-Language Models (RS-VLMs) necessitates a paradigm shift from recognition-centric perception to high-level deductive reasoning. 
The evolution of Remote Sensing Vision-Language Models(RS-VLMs) emphasizes the importance of transitioning from perception-centric recognition toward high-level deductive reasoning to enhance cognitive reliability in complex spatial tasks. However, current models often suffer from logical hallucinations, where correct answers are derived from flawed reasoning chains or rely on positional shortcuts rather than spatial logic. This decoupling undermines reliability in strategic spatial decision-making. To address this, we present GeoReason, a framework designed to synchronize internal thinking with final decisions. We first construct GeoReason-Bench, a logic-driven dataset containing 4,000 reasoning trajectories synthesized from geometric primitives and expert knowledge. We then formulate a two-stage training strategy: (1) Supervised Knowledge Initialization to equip the model with reasoning syntax and domain expertise, and (2) Consistency-Aware Reinforcement Learning to refine deductive reliability. This second stage integrates a novel Logical Consistency Reward, which penalizes logical drift via an option permutation strategy to anchor decisions in verifiable reasoning traces. Experimental results demonstrate that our framework significantly enhances the cognitive reliability and interpretability of RS-VLMs, achieving state-of-the-art performance compared to other advanced methods.
\end{abstract}


\section{Introduction}
\label{sec:intro}
\begin{figure}
    \centering
    \includegraphics[width=0.85\linewidth]{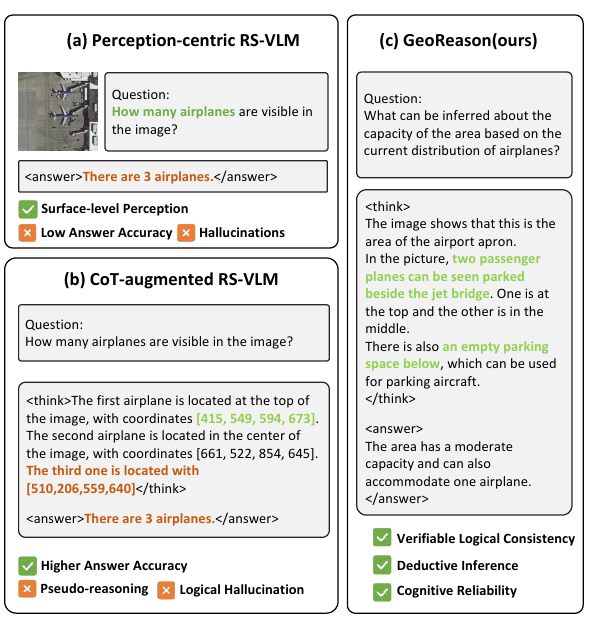}
    \caption{RS-VLM paradigm evolution. (a) \textbf{Perception-centric}: limited to surface-level identification with low accuracy. (b) \textbf{CoT-augmented}: improved accuracy but prone to pseudo-reasoning and logical decoupling. (c) \textbf{GeoReason}: achieves verifiable logical consistency and cognitive reliability via consistency-aware RL.}
    \label{fig:placeholder}
    \vspace{-6mm}
\end{figure}

Large foundation models have enabled Remote Sensing Vision-Language Models (RS-VLMs) to transition from traditional perception tasks including classification, detection, and segmentation toward comprehensive and interactive scene interpretation~\cite{hu2025rsgpt,li2024vision,zhan2025skyeyegpt,wen2026d3rdetrdetrdualdomaindensity}. Despite substantial success in tasks such as object identification~\cite{shen2025vlm,wen2025fanet,xiang2025infrared} and basic visual question answering (VQA)~\cite{lin2025rs}, current models encounter critical bottlenecks in complex cognitive scenarios~\cite{ding2025rethinking}. As illustrated in Fig. 1(a)(b), perception-centric models are prone to visual hallucinations, while Chain-of-Thought (CoT)~\cite{wei2022chain} prompting often introduces ``pseudo-reasoning''—a phenomenon where correct conclusions are derived from flawed logic or positional shortcuts rather than spatial evidence~\cite{bennett1994spatial,koehler1989veridical}. Such decoupling between reasoning trajectories and factual evidence severely undermines the cognitive reliability of RS-VLMs in strategic decision-making tasks, such as capacity estimation or functional zoning~\cite{muhtar2025quality}.

To bridge this gap, we propose \textbf{GeoReason}, a framework designed to align internal reasoning with final decisions through a consistency-aware pipeline (Fig. 1(c)). Our contribution is twofold: 
1) We curate \textbf{GeoReason-Bench}, a logic-driven dataset of 4,000 high-fidelity reasoning trajectories synthesized from geometric primitives and expert-knowledge pipelines that transform morphological patterns into verifiable logic. 
2) We develop a two-stage training strategy consisting of \textbf{Supervised Knowledge Initialization}~\cite{chu2025sft} followed by \textbf{Consistency-Aware Reinforcement Learning}~\cite{chen2025grpo}. By leveraging Group Relative Policy Optimization (GRPO)~\cite{shao2024deepseekmath} with a novel \textbf{Logical Consistency Reward (LCR)}, our method employs an option permutation strategy to penalize logical drift. This approach compels the model to internalize sound deductive derivation, effectively bridging the gap between perceptual recognition and high-level deductive in remote sensing~\cite{zhou2025scientists}.

\section{Proposed Methodology}
\label{sec:method}
\begin{figure*}
    \centering
    \includegraphics[scale=0.7]{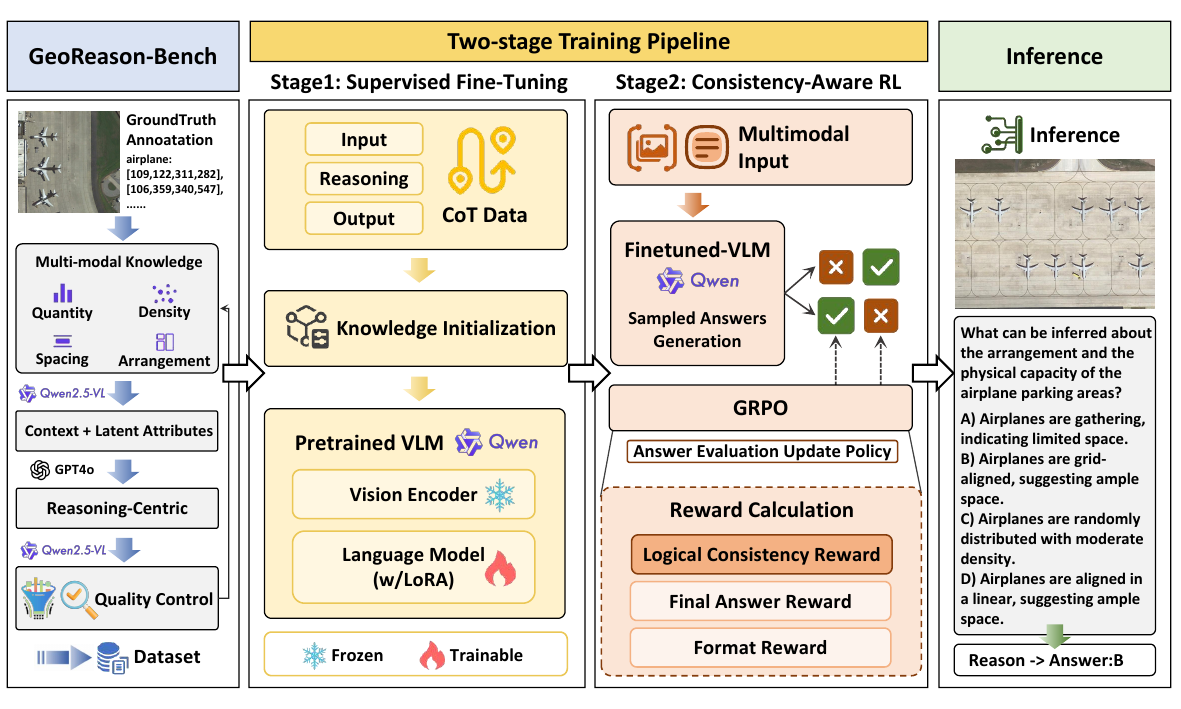}
    \vspace{-0.3cm}
    \caption{Overview of the GeoReason framework: (Left) logic-driven curation of GeoReason-Bench via multimodal knowledge integration; (Middle) two-stage training pipeline comprising Supervised Fine-Tuning and Consistency-Aware Reinforcement Learning; (Right) the resulting deductive inference process.}
    \label{fig:overview}
    \vspace{-0.4cm}
\end{figure*}

The overall architecture of GeoReason is illustrated in Fig.~\ref{fig:overview}. It bridges the gap between raw perception and cognitive reasoning through a logic-driven curation pipeline and a consistency-reinforced training strategy.

\subsection{GeoReason-Bench: A Logic-Driven Curation Pipeline}
To transition from perception-centric tasks to high-level cognitive interpretation, we develop a pipeline that transforms raw geometric primitives into high-fidelity reasoning trajectories. This process is structured into two primary phases: multi-modal knowledge integration and logic-augmented synthesis.

\subsubsection{Multi-modal Knowledge Integration} 
We first derive domain-specific structural features from the DOTA~\cite{xia2018dota} and DIOR~\cite{li2020object} repositories. Beyond standard bounding boxes, we extract geometric primitives (e.g., scale and orientation) and aggregate them into morphological patterns, such as spatial density, inter-object spacing, and clustering configurations (e.g., grid-like residential zones vs. linear logistics hubs)~\cite{xiang2026slgnetsynergizingstructuralpriors}. To bridge the gap between these discrete features and high-level reasoning, we further employ a state-of-the-art VLM to generate holistic scene descriptions, capturing global context and latent environmental attributes. By synthesizing these natural language narratives with structural metadata, we create a multi-layered representation that ensures the subsequent reasoning generation is grounded in both precise geometric priors and rich semantic context.

\subsubsection{Logic-Augmented Synthesis and Quality Control} The integrated features are serialized into structured prompts for GPT-4o to synthesize reasoning-centric samples, each consisting of a Reasoning Trajectory $\mathcal{T}$ (Chain-of-Thought) and a Final Answer $\mathcal{A}$. The dataset is stratified into two functional subsets: a Perception-Logic Subset ($D_{SFT}$, 1k) focusing on multi-step spatial integration, and a Deductive-Reasoning Subset ($D_{RL}$, 3k) formatted as Multiple-Choice Questions (MCQs) targeting high-level challenges like functional zoning and capacity estimation. To ensure logical integrity, we implement a dual-gate quality control mechanism. First, a cross-model consistency check is performed using a secondary VLM to prune "logical hallucinations" where the reasoning contradicts the visual evidence. Second, a manual expert review is conducted on a 10\% representative sample to calibrate linguistic precision and verify the domain logic. This rigorous refinement process ensures that GeoReason-Bench provides a high-fidelity foundation for the subsequent reinforcement learning stage.

\subsection{Two-Stage training pipeline}
To internalize the deductive capabilities of GeoReason-Bench, we develop a two-stage training pipeline (Fig.~\ref{fig:overview} middle). Stage 1 employs Supervised Knowledge Initialization to equip the model with foundational reasoning syntax and domain expertise. Stage 2 implements Consistency-Aware Reinforcement Learning via Group Relative Policy Optimization (GRPO) to drive the transition from supervised imitation to autonomous logical correction.

\subsubsection{Supervised Knowledge Initialization} The first stage involves Supervised Fine-Tuning (SFT), a process primarily focuses on equipping the model with the fundamental syntax of Chain-of-Thought (CoT) and remote sensing domain expertise. By utilizing the perception-logic subset $D_{SFT}$, we fine-tune the initial model to minimize the standard auto-regressive cross-entropy loss:
\begin{equation}
\mathcal{L}_{SFT} = -\sum_{t=1}^T \log P(y_t | y_{<t}, \mathcal{I}, \mathcal{X})
\end{equation}
where $\mathcal{I}$ is the input image, $\mathcal{X}$ denotes the structured prompt, and $y$ represents the target sequence comprising the reasoning trajectory $\mathcal{T}$ and the final answer $\mathcal{A}$. Through SFT, the model learns to bridge raw visual features with linguistically coherent and spatially grounded reasoning chains, providing a stable policy initialization for the subsequent reinforcement learning phase.

\subsubsection{Consistency-Aware Reinforcement Learning}
% \begin{figure}
%     \centering
%     \includegraphics[width=\linewidth]{Fig_RS/LCR.pdf}
%     \caption{Caption}
%     \label{fig:placeholder}
%     \vspace{-4mm}
% \end{figure}
Building upon the SFT initialization, we employ Group Relative Policy Optimization (GRPO)~\cite{shao2025asking} to further refine the model's deductive reliability. GRPO is particularly suited for remote sensing tasks as it eliminates the memory-intensive critic network by utilizing group-level relative rewards to estimate the baseline. For each input, we sample a group of $G$ outputs $\{o_1, o_2, ..., o_G\}$, and the total reward assigned to each trajectory $i$ is formulated as:
\begin{equation}
R_i = r_{acc} + r_{fmt} + r_{LCR}
\end{equation}
where $r_{acc}$ and $r_{fmt}$ denote the rewards for outcome accuracy and format compliance, respectively. 

To specifically mitigate the "logical hallucination" problem, we propose a novel \textbf{Logical Consistency Reward (LCR)}, denoted as $r_{LCR}$. This mechanism utilizes an \textbf{Option Permutation} strategy to ensure that the model's decision is strictly anchored in its reasoning trace rather than positional shortcuts. Given an image $\mathcal{I}$ and query $q$, the model generates a reasoning trace $t$ and an initial answer $a$. We then apply a permutation $\mathcal{P}(\cdot)$ to shuffle the options, yielding a rephrased query $q' = \mathcal{P}(q)$. A "frozen-logic" second pass is performed to predict a secondary answer $\tilde{a} \sim \pi_{\theta}(\cdot | \mathcal{I}, q', t)$. The $r_{LCR}$ is then designed to penalize \textit{logical drift} where the conclusion shifts despite an identical reasoning trace:
\begin{equation}
r_{LCR} = \ln(e + L_t) \cdot \Phi(a, \tilde{a}) - \Omega(a, \tilde{a})
\end{equation}
where $L_t$ is the trace length. The core logic $\Phi$ grants a bonus $\alpha$ if both $a$ and $\tilde{a}$ are correct and semantically consistent, while $\Omega$ imposes a penalty $\eta$ if $a$ and $\tilde{a}$ lead to contradictory conclusions.

The policy is updated by maximizing the following objective function:
\begin{equation}
\mathcal{J}(\theta) = \mathbb{E} \left[ \frac{1}{G} \sum_{i=1}^{G} \mathcal{L}_i - \beta \mathbb{D}_{\text{KL}}(\pi_{\theta} || \pi_{\text{ref}}) \right]
\end{equation}
where the group-wise clipped loss $\mathcal{L}_i$ is defined as:
\begin{equation}
\mathcal{L}_i = \min(w_i A_i, \text{clip}(w_i, 1-\epsilon, 1+\epsilon) A_i)
\end{equation}
The importance weight $w_i$ and normalized advantage $A_i$ are given by:
\begin{equation}
w_i = \frac{\pi_{\theta}(o_i | q)}{\pi_{\theta_{\text{old}}}(o_i | q)}, \quad A_i = \frac{R_i - \text{mean}(R)}{\text{std}(R)}
\end{equation}
This reinforcement learning stage compels the model to internalize expert-level decision-making through sound logical derivation rather than stochastic guessing.
\section{EXPERIMENT}
\subsection{Experimental Setup}
\subsubsection{Datasets}
We conduct experiments on GeoReason-Bench, a logic-driven dataset containing 4,000 high-fidelity reasoning trajectories. It comprises two subsets: a Perception-Logic Subset ($D_{SFT}$, 1k), and a Deductive-Reasoning Subset ($D_{RL}$, 3k).
\subsubsection{Evaluation Metrics}
To provide a comprehensive assessment, we utilize three quantitative metrics:
\begin{itemize}
    \item Per-category Accuracy: Five distinct task dimensions: count, color, shape, reason, and scene (rural or urban).
    \item Overall Accuracy(OA): The ratio of the total number of correctly predicted samples to the total size of the test set.
    \item Average Accuracy(AA): The mean of the accuracies achieved across the five categories.
\end{itemize}
\subsubsection{Implementation Details}
We utilize Qwen2.5-VL-7B ~\cite{bai2025qwen2} as the base model with LoRA (rank 16). The training follows a two-stage process: 1) SFT for 1 epoch with a learning rate (LR) of $1 \times 10^{-4}$; 2) GRPO for 1200 steps with an LR of $1 \times 10^{-6}$.
\begin{table*}[!tb]
    \centering
    \footnotesize
    \begin{minipage}{0.7\textwidth} 
    
        \caption{Comparative analysis of our proposed GeoReason against state-of-the-art baselines on perceptual and reasoning tasks. The best results are highlighted in bold, and the second-best results are underlined.}
        \label{tab:main_result}
        \vspace{-0.5em}
        
        \resizebox{\textwidth}{!}{%
            \setlength{\tabcolsep}{5pt} 
            \begin{tabular}{lccccccc} 
                \toprule
                \multirow{2}{*}{\textbf{Method}} & \multicolumn{4}{c}{\textbf{Perceptual Tasks}} & \textbf{Reasoning} & \multicolumn{2}{c}{\textbf{Overall Metrics}} \\
                \cmidrule(r){2-5} \cmidrule(lr){6-6} \cmidrule(l){7-8}
                 & Count & Color & Shape & Scene & Reason & OA & AA \\
                \midrule
                
                \multicolumn{8}{l}{\textit{\color{gray}Close-source Commercial Vision-Language Models}} \\ 
                \addlinespace[0.2em]
                % Reason: 33.68 是次优 (第一是 43.51)
                GPT-4o~\cite{schulman2022chatgpt} & 6.35 & 42.27 & 34.57 & 92.06 & \underline{33.68} & 38.54 & 41.79 \\
                \midrule
                
                \multicolumn{8}{l}{\textit{\color{gray}Open-source Vision-Language Models}} \\ 
                \addlinespace[0.2em]
                Llava~\cite{li2023llava} & 9.52 & 44.33 & 18.52 & 90.48 & 16.84 & 28.69 & 35.94 \\
                % Count: 25.40 是次优 (第一是 34.92)
                % AA: 43.40 是次优 (第一是 56.20)
                Qwen2.5-VL~\cite{bai2025qwen2} & \underline{25.40} & 40.21 & 37.04 & 90.48 & 23.86 & 35.65 & \underline{43.40} \\
                \midrule
        
                \multicolumn{8}{l}{\textit{\color{gray}Open-source Remote Sensing Vision-Language Models}} \\ 
                \addlinespace[0.2em]
                % Color: 45.36 是次优 (第一是 56.70)
                % Shape: 45.68 是次优 (第一是 50.62)
                % OA: 38.71 是次优 (第一是 51.27)
                RS-EoT~\cite{shao2025asking} & 19.05 & \underline{45.36} & \underline{45.68} & 68.25 & 32.28 & \underline{38.71} & 42.12 \\
                GeoChat~\cite{kuckreja2024geochat} & 12.70 & 17.53 & 20.99 & 90.48 & 16.49 & 24.79 & 31.64 \\
                % Scene: 93.65 是次优 (第一是 95.23)
                SkySenseGPT~\cite{luo2024skysensegpt} & 14.29 & 24.74 & 23.46 & \underline{93.65} & 14.39 & 25.81 & 34.10 \\
                \midrule
                
                \rowcolor{gray!15} 
                \textbf{GeoReason(Ours)} & \textbf{34.92} & \textbf{56.70} & \textbf{50.62} & \textbf{95.23} & \textbf{43.51} & \textbf{51.27} & \textbf{56.20} \\
                
                \bottomrule
            \end{tabular}
        }
    \end{minipage}
\end{table*}

\subsection{Quantitative Results}
Table~\ref{tab:main_result} summarizes the quantitative performance on the GeoReason-Bench test set. We evaluate models across Perceptual Tasks (Count, Color, Shape, Scene) and Reasoning Tasks (Reason) to analyze their multi-level understanding. As shown in Table~\ref{tab:main_result}, GeoReason significantly outperforms all baselines, achieving an Overall Accuracy (OA) of 51.27\% and an Average Accuracy (AA) of 56.20\%. Notably, in the Reasoning task, our framework achieves 43.51\%, surpassing the base Qwen2.5-VL by 19.65\% and the commercial GPT baseline by 9.83\%. These results demonstrate that our consistency-aware reinforcement learning effectively bridges the gap between surface-level perception and high-level deductive inference, compelling the model to anchor its decisions in verifiable spatial logic rather than stochastic guessing.

\begin{table}[!tb]
    \centering
    \caption{Ablation study of different training stages on the GeoReason-Bench test set.}
    \vspace{-0.5em}
    \setlength{\tabcolsep}{3pt}
    \begin{tabular}{lcc}
        \toprule
        \textbf{Configuration} & \textbf{AA} (\%) & \textbf{Reason Acc} (\%) \\
        \midrule
        Base Model (Qwen2.5-VL) & 43.40 & 23.86 \\
        + SFT & 47.92 & 31.93 \\
        + GRPO (Standard) & 53.18 & 36.49 \\
         \rowcolor{gray!15}\textbf{+ GRPO (Consistency Reward)} & \textbf{56.20} & \textbf{43.51} \\
        \bottomrule
    \end{tabular}
    \label{tab:ablation}
    \vspace{-2mm}
\end{table}

\subsection{Ablation Study}
The ablation study in Table~\ref{tab:ablation} illustrates the incremental impact of each component on mitigating logical hallucinations. The SFT stage establishes the basic ``Reason-Answer'' paradigm, elevating Reasoning Accuracy to 31.93\%. While the subsequent integration of standard GRPO improves Average Accuracy to 53.81\%, the Reasoning Accuracy lags at 36.49\%. This gap underscores a critical process-outcome misalignment, where the model prioritizes visual shortcuts over genuine deduction. The proposed Logical Consistency Reward (LCR) successfully bridges this gap, driving Reasoning Accuracy to 43.51\%. By penalizing logical contradictions, LCR compels the model to treat the reasoning chain as essential evidence, effectively suppressing hallucinations and ensuring that final answers are anchored in verifiable spatial logic.

\begin{figure}
    \centering
    \includegraphics[width=\linewidth]{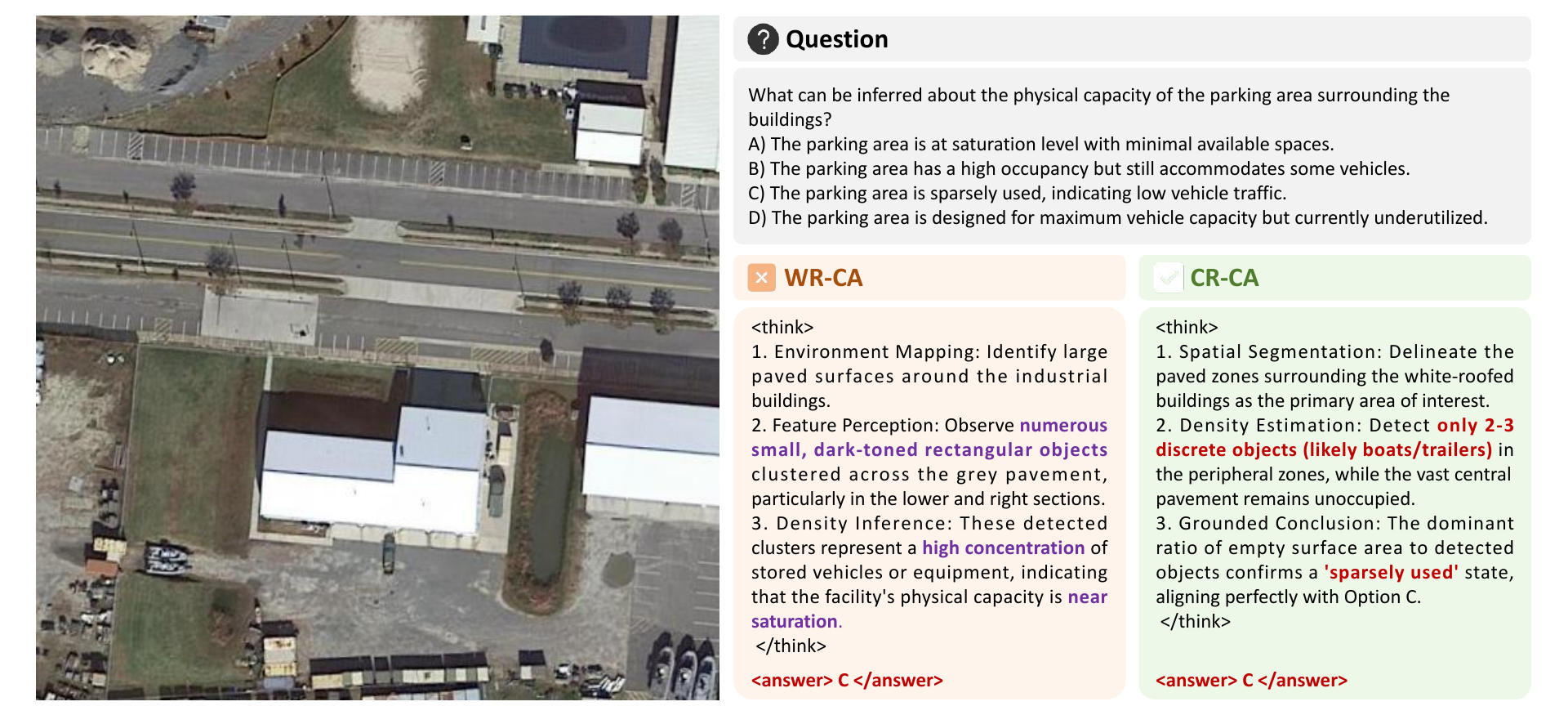}
    \vspace{-6mm}
    \caption{Reasoning-Answer alignment comparison. \textcolor{red}{Red} denotes correct answer or reasoning trace and \textcolor{purple}{purple} denotes flawed reasoning components. The left case is an example of \textbf{Wrong Reasoning but Correct Answer (WR-CA)}, indicating logical hallucination, while the right case is an example of \textbf{Correct Reasoning and Correct Answer (CR-CA)}, demonstrating logically sound deductive interpretation.}
    \label{fig:placeholder}
    \vspace{-6mm}
\end{figure}

\subsection{Qualitative Analysis}
Fig.~\ref{fig:placeholder} illustrates the impact of LCR on a representative parking area utilization task. The standard GRPO baseline (WR-CA) exhibits severe logic hallucination; it paradoxically claims the area is ``near saturation'' with ``numerous objects'' while selecting Option C (\textit{Sparsely used}). In contrast, GeoReason (CR-CA) demonstrates robust evidential support by accurately identifying the ``dominant empty surface'' and ``2-3 discrete objects''. This alignment confirms that LCR effectively forces the model to ground its reasoning in visual evidence, eliminating the reliance on spurious correlations.

\section{Conclusion}
In this paper, we presented GeoReason, a novel framework designed to mitigate logical hallucinations and pseudo-reasoning in Remote Sensing Vision-Language Models (RS-VLMs). By introducing GeoReason-Bench, we provided a high-fidelity foundation of 4,000 reasoning trajectories that transform geometric primitives into structured deductive logic. Our two-stage training pipeline, which integrates Group Relative Policy Optimization (GRPO) with a specialized Logical Consistency Reward (LCR), compels the model to anchor its final decisions strictly within verifiable reasoning traces. Experimental results demonstrate that GeoReason significantly enhances both overall accuracy and cognitive reliability, effectively ensuring that the model provides the right answers for the right reasons.

\small
\bibliographystyle{IEEEtranN}
\bibliography{main}

% Generated by IEEEtranN.bst, version: 1.14 (2015/08/26)
\begin{thebibliography}{26}
\providecommand{\natexlab}[1]{#1}
\providecommand{\url}[1]{#1}
\csname url@samestyle\endcsname
\providecommand{\newblock}{\relax}
\providecommand{\bibinfo}[2]{#2}
\providecommand{\BIBentrySTDinterwordspacing}{\spaceskip=0pt\relax}
\providecommand{\BIBentryALTinterwordstretchfactor}{4}
\providecommand{\BIBentryALTinterwordspacing}{\spaceskip=\fontdimen2\font plus
\BIBentryALTinterwordstretchfactor\fontdimen3\font minus \fontdimen4\font\relax}
\providecommand{\BIBforeignlanguage}[2]{{%
\expandafter\ifx\csname l@#1\endcsname\relax
\typeout{** WARNING: IEEEtranN.bst: No hyphenation pattern has been}%
\typeout{** loaded for the language `#1'. Using the pattern for}%
\typeout{** the default language instead.}%
\else
\language=\csname l@#1\endcsname
\fi
#2}}
\providecommand{\BIBdecl}{\relax}
\BIBdecl

\bibitem[Hu et~al.(2025)Hu, Yuan, Wen, Lu, Liu, and Li]{hu2025rsgpt}
Y.~Hu, J.~Yuan, C.~Wen, X.~Lu, Y.~Liu, and X.~Li, ``Rsgpt: A remote sensing vision language model and benchmark,'' \emph{ISPRS Journal of Photogrammetry and Remote Sensing}, vol. 224, pp. 272--286, 2025.

\bibitem[Li et~al.(2024)Li, Wen, Hu, Yuan, and Zhu]{li2024vision}
X.~Li, C.~Wen, Y.~Hu, Z.~Yuan, and X.~X. Zhu, ``Vision-language models in remote sensing: Current progress and future trends,'' \emph{IEEE Geoscience and Remote Sensing Magazine}, vol.~12, no.~2, pp. 32--66, 2024.

\bibitem[Zhan et~al.(2025)Zhan, Xiong, and Yuan]{zhan2025skyeyegpt}
Y.~Zhan, Z.~Xiong, and Y.~Yuan, ``Skyeyegpt: Unifying remote sensing vision-language tasks via instruction tuning with large language model,'' \emph{ISPRS Journal of Photogrammetry and Remote Sensing}, vol. 221, pp. 64--77, 2025.

\bibitem[Wen et~al.(2026)Wen, Yang, Bao, Zhang, Xiang, Li, and Liu]{wen2026d3rdetrdetrdualdomaindensity}
\BIBentryALTinterwordspacing
Z.~Wen, Z.~Yang, X.~Bao, L.~Zhang, X.~Xiang, W.~Li, and Y.~Liu, ``D$^3$r-detr: Detr with dual-domain density refinement for tiny object detection in aerial images,'' 2026. [Online]. Available: \url{https://arxiv.org/abs/2601.02747}
\BIBentrySTDinterwordspacing

\bibitem[Shen et~al.(2025)Shen, Liu, Li, Fang, Ma, Liao, Shen, Zhang, Zhao, Zhang, et~al.]{shen2025vlm}
H.~Shen, P.~Liu, J.~Li, C.~Fang, Y.~Ma, J.~Liao, Q.~Shen, Z.~Zhang, K.~Zhao, Q.~Zhang \emph{et~al.}, ``Vlm-r1: A stable and generalizable r1-style large vision-language model,'' \emph{arXiv preprint arXiv:2504.07615}, 2025.

\bibitem[Wen et~al.(2025)Wen, Li, Liu, Chen, Xiang, Li, Wang, Zhao, and Zhou]{wen2025fanet}
Z.~Wen, P.~Li, Y.~Liu, J.~Chen, X.~Xiang, Y.~Li, H.~Wang, Y.~Zhao, and G.~Zhou, ``Fanet: Frequency-aware attention-based tiny-object detection in remote sensing images,'' \emph{Remote Sensing}, 2025.

\bibitem[Xiang et~al.(2025)Xiang, Zhou, Niu, Pan, Huang, Li, Wen, Qi, and Gao]{xiang2025infrared}
X.~Xiang, G.~Zhou, B.~Niu, Z.~Pan, L.~Huang, W.~Li, Z.~Wen, J.~Qi, and W.~Gao, ``Infrared-visible image fusion meets object detection: Towards unified optimization for multimodal perception,'' \emph{Remote Sensing}, vol.~17, no.~21, p. 3637, 2025.

\bibitem[Lin et~al.(2025)Lin, Hong, Ge, Luo, Jiang, Jin, and Wen]{lin2025rs}
H.~Lin, D.~Hong, S.~Ge, C.~Luo, K.~Jiang, H.~Jin, and C.~Wen, ``Rs-moe: A vision-language model with mixture of experts for remote sensing image captioning and visual question answering,'' \emph{IEEE Transactions on Geoscience and Remote Sensing}, 2025.

\bibitem[Ding et~al.(2025)Ding, Li, Cao, and Shao]{ding2025rethinking}
Y.~Ding, L.~Li, B.~Cao, and J.~Shao, ``Rethinking bottlenecks in safety fine-tuning of vision language models,'' \emph{arXiv preprint arXiv:2501.18533}, 2025.

\bibitem[Wei et~al.(2022)Wei, Wang, Schuurmans, Bosma, Xia, Chi, Le, Zhou, et~al.]{wei2022chain}
J.~Wei, X.~Wang, D.~Schuurmans, M.~Bosma, F.~Xia, E.~Chi, Q.~V. Le, D.~Zhou \emph{et~al.}, ``Chain-of-thought prompting elicits reasoning in large language models,'' \emph{Advances in neural information processing systems}, vol.~35, pp. 24\,824--24\,837, 2022.

\bibitem[Bennett(1994)]{bennett1994spatial}
B.~Bennett, ``Spatial reasoning with propositional logics,'' in \emph{Principles of knowledge representation and reasoning}.\hskip 1em plus 0.5em minus 0.4em\relax Elsevier, 1994, pp. 51--62.

\bibitem[Koehler and Shaviro(1989)]{koehler1989veridical}
J.~J. Koehler and D.~N. Shaviro, ``Veridical verdicts: Increasing verdict accuracy through the use of overtly probabilistic evidence and methods,'' \emph{Cornell L. Rev.}, vol.~75, p. 246, 1989.

\bibitem[Muhtar et~al.(2025)Muhtar, Zhang, Li, Gu, He, Xiao, and Zhang]{muhtar2025quality}
D.~Muhtar, E.~Zhang, Z.~Li, F.~Gu, Y.~He, P.~Xiao, and X.~Zhang, ``Quality-driven curation of remote sensing vision-language data via learned scoring models,'' \emph{arXiv preprint arXiv:2503.00743}, 2025.

\bibitem[Chu et~al.(2025)Chu, Zhai, Yang, Tong, Xie, Schuurmans, Le, Levine, and Ma]{chu2025sft}
T.~Chu, Y.~Zhai, J.~Yang, S.~Tong, S.~Xie, D.~Schuurmans, Q.~V. Le, S.~Levine, and Y.~Ma, ``Sft memorizes, rl generalizes: A comparative study of foundation model post-training,'' \emph{arXiv preprint arXiv:2501.17161}, 2025.

\bibitem[Chen et~al.(2025)Chen, Ge, Wang, Ge, Cheng, Shan, and Liu]{chen2025grpo}
Y.~Chen, Y.~Ge, R.~Wang, Y.~Ge, J.~Cheng, Y.~Shan, and X.~Liu, ``Grpo-care: Consistency-aware reinforcement learning for multimodal reasoning,'' \emph{arXiv preprint arXiv:2506.16141}, 2025.

\bibitem[Shao et~al.(2024)Shao, Wang, Zhu, Xu, Song, Bi, Zhang, Zhang, Li, Wu, et~al.]{shao2024deepseekmath}
Z.~Shao, P.~Wang, Q.~Zhu, R.~Xu, J.~Song, X.~Bi, H.~Zhang, M.~Zhang, Y.~Li, Y.~Wu \emph{et~al.}, ``Deepseekmath: Pushing the limits of mathematical reasoning in open language models,'' \emph{arXiv preprint arXiv:2402.03300}, 2024.

\bibitem[Zhou et~al.(2025)Zhou, Wang, He, Shen, Xiao, Li, Feng, Guo, Yang, Wu, et~al.]{zhou2025scientists}
Y.~Zhou, Y.~Wang, X.~He, A.~Shen, R.~Xiao, Z.~Li, Q.~Feng, Z.~Guo, Y.~Yang, H.~Wu \emph{et~al.}, ``Scientists' first exam: Probing cognitive abilities of mllm via perception, understanding, and reasoning,'' \emph{arXiv preprint arXiv:2506.10521}, 2025.

\bibitem[Xia et~al.(2018)Xia, Bai, Ding, Zhu, Belongie, Luo, Datcu, Pelillo, and Zhang]{xia2018dota}
G.-S. Xia, X.~Bai, J.~Ding, Z.~Zhu, S.~Belongie, J.~Luo, M.~Datcu, M.~Pelillo, and L.~Zhang, ``Dota: A large-scale dataset for object detection in aerial images,'' in \emph{Proceedings of the IEEE conference on computer vision and pattern recognition}, 2018, pp. 3974--3983.

\bibitem[Li et~al.(2020)Li, Wan, Cheng, Meng, and Han]{li2020object}
K.~Li, G.~Wan, G.~Cheng, L.~Meng, and J.~Han, ``Object detection in optical remote sensing images: A survey and a new benchmark,'' \emph{ISPRS journal of photogrammetry and remote sensing}, vol. 159, pp. 296--307, 2020.

\bibitem[Xiang et~al.(2026)Xiang, Zhou, Wen, Li, Niu, Wang, Huang, Wang, Liu, Pan, and Hu]{xiang2026slgnetsynergizingstructuralpriors}
\BIBentryALTinterwordspacing
X.~Xiang, G.~Zhou, Z.~Wen, W.~Li, B.~Niu, F.~Wang, L.~Huang, Q.~Wang, Y.~Liu, Z.~Pan, and Y.~Hu, ``Slgnet: Synergizing structural priors and language-guided modulation for multimodal object detection,'' 2026. [Online]. Available: \url{https://arxiv.org/abs/2601.02249}
\BIBentrySTDinterwordspacing

\bibitem[Shao et~al.(2025)Shao, Li, Zhang, Xu, He, Yuan, He, Dai, Yan, Chen, et~al.]{shao2025asking}
R.~Shao, Z.~Li, Z.~Zhang, L.~Xu, X.~He, H.~Yuan, B.~He, Y.~Dai, Y.~Yan, Y.~Chen \emph{et~al.}, ``Asking like socrates: Socrates helps vlms understand remote sensing images,'' \emph{arXiv preprint arXiv:2511.22396}, 2025.

\bibitem[Bai et~al.(2025)Bai, Chen, Liu, Wang, Ge, Song, Dang, Wang, Wang, Tang, et~al.]{bai2025qwen2}
S.~Bai, K.~Chen, X.~Liu, J.~Wang, W.~Ge, S.~Song, K.~Dang, P.~Wang, S.~Wang, J.~Tang \emph{et~al.}, ``Qwen2. 5-vl technical report,'' \emph{arXiv preprint arXiv:2502.13923}, 2025.

\bibitem[Schulman et~al.(2022)Schulman, Zoph, Kim, Hilton, Menick, Weng, Uribe, Fedus, Metz, Pokorny, et~al.]{schulman2022chatgpt}
J.~Schulman, B.~Zoph, C.~Kim, J.~Hilton, J.~Menick, J.~Weng, J.~F.~C. Uribe, L.~Fedus, L.~Metz, M.~Pokorny \emph{et~al.}, ``Chatgpt: Optimizing language models for dialogue,'' \emph{OpenAI blog}, vol.~2, no.~4, 2022.

\bibitem[Li et~al.(2023)Li, Wong, Zhang, Usuyama, Liu, Yang, Naumann, Poon, and Gao]{li2023llava}
C.~Li, C.~Wong, S.~Zhang, N.~Usuyama, H.~Liu, J.~Yang, T.~Naumann, H.~Poon, and J.~Gao, ``Llava-med: Training a large language-and-vision assistant for biomedicine in one day,'' \emph{Advances in Neural Information Processing Systems}, vol.~36, pp. 28\,541--28\,564, 2023.

\bibitem[Kuckreja et~al.(2024)Kuckreja, Danish, Naseer, Das, Khan, and Khan]{kuckreja2024geochat}
K.~Kuckreja, M.~S. Danish, M.~Naseer, A.~Das, S.~Khan, and F.~S. Khan, ``Geochat: Grounded large vision-language model for remote sensing,'' in \emph{Proceedings of the IEEE/CVF Conference on Computer Vision and Pattern Recognition}, 2024, pp. 27\,831--27\,840.

\bibitem[Luo et~al.(2024)Luo, Pang, Zhang, Wang, Wang, Dang, Lao, Wang, Chen, Tan, et~al.]{luo2024skysensegpt}
J.~Luo, Z.~Pang, Y.~Zhang, T.~Wang, L.~Wang, B.~Dang, J.~Lao, J.~Wang, J.~Chen, Y.~Tan \emph{et~al.}, ``Skysensegpt: A fine-grained instruction tuning dataset and model for remote sensing vision-language understanding,'' \emph{arXiv preprint arXiv:2406.10100}, 2024.

\end{thebibliography}

\end{document}